# Structured learning and detailed interpretation of minimal object images

Guy Ben-Yosef [1,2,3]    Liav Assif [2]    Shimon Ullamn [1,3]

[1] MIT Computer Science and Artificial Intelligence Laboratory
[2] Weizmann Institute, Department of Computer Science and Applied Mathematics
[3] MIT Centre for Brains, Minds and Machines
`gby@csail.mit.edu, liav.assif@weizmann.ac.il , shimon.ullman@weizmann.ac.il`

## Abstract

We model the process of human 'full interpretation' of object images, namely the ability to identify and localize all semantic features and parts that are recognized by human observers. The task is approached by dividing the interpretation of the complete object to the interpretation of multiple reduced but interpretable local regions. We model interpretation by a structured learning framework, in which there are primitive components and relations that play a useful role in local interpretation by humans. To identify useful components and relations used in the interpretation process, we consider the interpretation of 'minimal configurations', namely reduced local regions that are minimal in the sense that further reduction will turn them unrecognizable and uninterpretable. We show experimental results of our model, and results of predicting and testing relations that were useful to the model via transformed minimal images.

## 1. Image interpretation: detecting structure in image

In this work we aim to model human 'full interpretation' of object images, which is the ability to identify and localize all semantic features and parts that are recognized by human observers. The task is approached by dividing the interpretation of complete objects to interpretation of multiple local regions that are still interpretable to humans. In such reduced regions interpretation is simpler, since the number of semantic components is small, and the variability of possible configurations is low.

In our interpretation study we developed a model of human representation and learning of structure in object images, by studying the semantic components and their relations in local object regions, which are minimal recognizable and interpretable configurations. Our study contributes to the computer vision community in two ways: (i) it identifies features and principles for structural models coming from a human interpretation study. (ii) It attacks, perhaps for the first time, the problem of 'full' human-level interpretation of object images (as defined above), and provides a modeling approach for getting detailed interpretation of visual scenes.

Current approaches for object interpretation include FCNNs [1], which produce low-resolution semantic segments (due to their pooling layers), combined with structured prediction models such as dense CRF to achieve more accurate boundaries (e.g., [2-4]). Such models are able to detect large part segments [2] or keypoints [5-6], but their accuracy is lower when dealing e.g., with complex variable configurations [6], multiple parts [7], and parts from two objects that are mixed due to interactions between objects [8]. Possible reasons are (i) use of CRFs that are loopy graphical models in which optimization is difficult, and (ii) a focus on limited binary potentials rather than more complex and higher order relations [3,4,9]. Our human interpretation study provides alternative directions by (i) considering local interpretation in which complexity is reduced, and (ii) offering relations that we show to be important for human interpretation.

## 2. Interpretation study on minimal configurations

In performing local interpretation, how should an object image be divided into local regions? The approach we take in this study is to develop and test the interpretation model of regions that can be interpreted on their own by human observers, but at the same time are as limited as possible. We used for this a set of local recognizable images derived by a recent study on minimal recognizable images [10].

### 2.1. Minimal configurations

A 'minimal configuration' (also termed Minimal Recognizable Configuration, or MIRC) is defined as an image patch that can be reliably recognized by human observers, which is minimal in the sense that further reduction by either size or resolution makes the patch unrecognizable. A search started with images from different object classes, and identified their minimal configurations over all possible positions and sizes. Examples of minimal images are in Fig. 1A.

Two notable aspects of the psychophysics results were used in the interpretation study. The first is the presence of a sharp transition for almost all minimal configurations from a recognizable to a non-recognizable minimal image: a surprisingly small change at the minimal-configuration level can make it unrecognizable. Examples are shown in

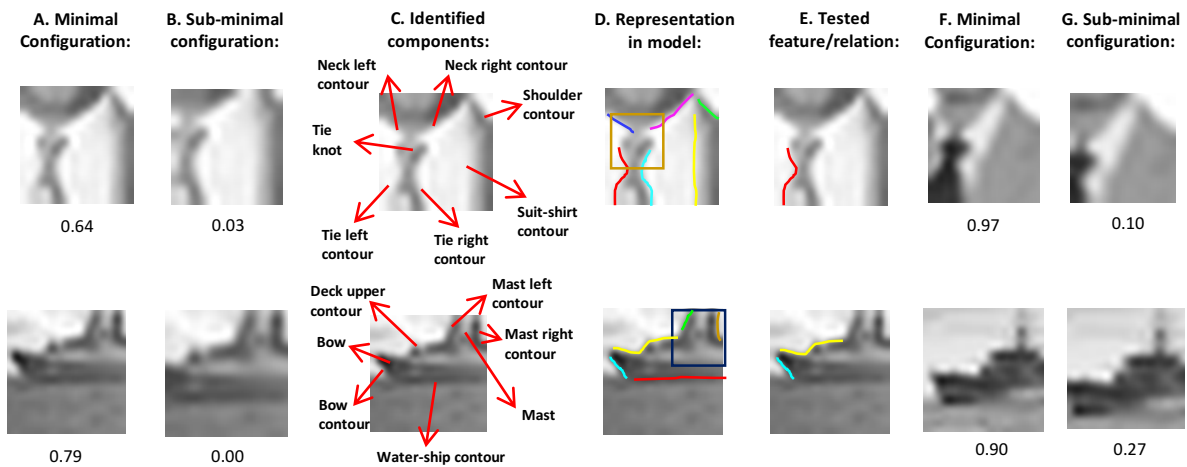

*Figure 1.* Inferring relations from minimal configurations. A minimal configuration (A) and its unrecognizable sub-minimal reduced version (B, recognition rate shown below the images), are shown with the internal components recognized by humans in the minimal images (C). The components in our model are represented with points, contours, and square regions (D). To identify useful components and relations for interpretation, we compared the minimal and sub-minimal images. Using the identified components, we found if any component in (A) are missing in (B). The contribution of each missing component or relation was then evaluated using training examples (see text). When necessary, several alternatives were evaluated. Examples of informative components and relations are shown in (E). Examples of additional minimal / sub-minimal pairs in the training set with the same missing component or relation, with its effect on recognition, are shown in (F-G). Inferred components and relations illustrated in the figure are missing contour element (top row), and connectedness of two contours at high curvature point (bottom row).

Fig. 1A-B, together with their respective recognition rates. The loss of recognition when the image is sufficiently reduced and features are removed is expected, but the sharp drop at the minimal level is remarkable, and consistent across many examples [10].

The second aspect is that humans could consistently recognize multiple semantic features and parts within the minimal images. It is concluded that recognition and interpretation go hand in hand, and that minimal recognizable configurations are also the minimal interpretable ones [9-10].

## 2.2. Interpretation model

Our interpretation scheme has two main components: in the learning stage, it learns the semantic structure of an image in a supervised manner, and in the interpretation (inference) stage, it identifies the learned structure in similar image regions.

**learning the semantic structure:** The semantic features to be identified by the model (e.g., 'ear', 'tie knot', etc.) were features that human observers label consistently in minimal images, verified using an MTurk procedure (the average number of consistently identified elements within a single minimal image was 8). The semantic features were then represented by three types of visual primitives: points (e.g., a horse eye, which can contain 1 or 2 pixels in minimal images), contours (for borders, e.g., a tie border), and square region primitives. See examples in Fig. 1C-D.

Given these semantic elements, we prepared a set of annotated images, in which the semantic components were marked manually on multiple examples of the minimal image, and then used in a structured learning framework based on a random forest classifier [9]. Our learning scheme computes a set of relations between elements in the structure for both positive and negative examples, and then learns the contribution of each relation to interpretation. A critical component in this scheme is therefore the types of relations that were used.

**Interpretation for novel image:** The interpretation process starts with a candidate region and its proposed category (e.g., that it contains a horse-head). The process then used the learned model of the region's structure to identify within the region a structure that best approximates the learned one. This process proceeds in two main stages. The first is a search for local primitives, namely points, contours, and regions in the image, to serve as potential candidates for the different components of the expected structure. The second stage searches for a configuration of the components that best matches the learned structure.

## 3. Useful types of relations

The model described in this work belongs to the general approach of structured models. There is a rich history to the use of structural models in the computational study of vision, including visual recognition and interpretation (e.g., [11-13,2-4]). Models differ in the shape components used to create structured configurations, the relations used to represent configurations (we consider properties and attributes of a single element as unary relations), and the algorithms used to learn structures from image examples, and to identify similar structure in novel images.

The relations used in these models were mostly simple, in particular, the expected location within a reference frame and relative displacement (e.g., [2,11], hereinafter the 'basic relations'), but a few used more complex relations such as co-termination [12], parallelism of elements [13], and containment [3].

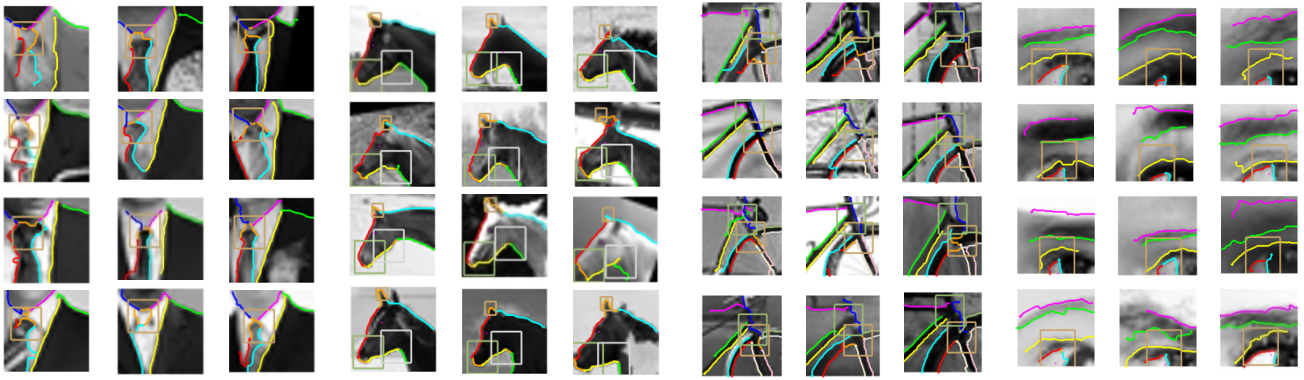

*Figure 2: Interpretation results for minimal images belonging to (left to right) a man in a suit, a horse-head, a bike, and an eye.*

In the human and primate vision literature there has also been a great deal of work on relations between elements in the visual field. These works have shown sensitivity of the visual system to known principles of perceptual organization such as proximity, similarity, connectivity, symmetry and continuity between visual elements, and also to parallelism, curvature, convexity, co-linearity, co-circularity, connectedness of contours, and inclusion between elements (see review in [9]).

The availability of minimal images (sec. 2.1) allowed us to examine whether local appearance and basic relations are sufficient for producing an accurate 'full' interpretation by our model. Minimal configurations are by construction non-redundant visual patterns, and therefore their recognition and interpretation depend on the effective use of all the available visual information. It consequently becomes of interest to examine the performance of a model that uses a limited set of relations when applied to the interpretation of minimal images, and compare to interpretation produced by a model with a richer set of relations.

In the case of minimal images, the sharp drop in human's ability to recognize and interpret a minimal configuration when the image is slightly reduced, provided a tool for identifying useful relations for modeling human interpretation. A minimal image was compared with its similar, but unrecognizable sub-image, to identify either a missing component (e.g., a contour, as in Fig. 1E, top row) or a relation (e.g., between two contours, as in Fig. 1E, bottom row), which were present in the minimal image but not in the sub-minimal configuration. For each considered component or relation, we tested its consistent effect on other pairs of minimal and sub-minimal images (Fig. 1F-G), and we evaluated its statistical contribution to the learning process, by adding it to the set of relations, training a new interpretation algorithm, and measuring the difference in interpretation performance with and without this relation. A list of the most contributive relations is shown in Table 1 (hereinafter, 'extended' relations set).

The interpretations produced by the model were compared with the ground truth annotations supplied by human annotators. To assess the role of the extended relations derived from minimal and sub-minimal pairs (Table 1), we compared results from two versions of our model, which differed in the relations included in the model: one using only the basic (relations 1,4,5 in Table 1), and the other using the extended set of relations in Table 1.

Fig. 2 shows examples of the interpretations produced by the model with the extended set for novel test images. To assess the interpretations, we matched the model output to human annotations for multiple examples. Our training set contained 120 positive examples, and 25,000 negative examples for each interpretation model. Our test set contained 480 examples for the horse-head minimal image, 330 examples for the man-in-suit minimal image, and 120 of the eye and the bike minimal images (Fig. 2). We automatically matched the ground truth annotated primitives to the interpretation output by the Jaccard overlap index (a.k.a., IoU). Our results show 0.48 accuracy in average for the basic set, 0.62 accuracy in average for the extended relations set, and 0.78 agreement between different human annotators, which served as an upper bound for interpretation. Interpretation using the extended set of relations was significantly closer to ground truth compared with the use of basic set of relations, but still far from human interpretation. More results and evaluation details are discussed in [9].

## 4. Intervention on minimal images

The interpretation model includes informative relations between components which were identified using the dataset of sub-minimal images. The model predicts that disrupting these relations should reduce the ability of human observers to recognize and interpret minimal images. To further verify the role of these relations, we used direct intervention on minimal images, testing whether removing specific relations from the minimal image will decrease human recognition. For this purpose we created transformed versions of the minimal images (e.g., rendering sketches, and re-coloring small set of pixels), in which specific relations were selectivity manipulated. The transformed versions were then tested psychophysically via the MTurk. The tested relations for each minimal image were taken from the most informative relations predicted by its interpretation model, for example, the minimum intensity property of the horse eye (Fig. 3A), or the straightness and parallelism relations of the bike tubes' contours (Fig. 3B). For many of the tested relations, the manipulation resulted in a significant drop in recognition rate [9], which further support the role of complex relations in human interpretation process. We conclude that computer vision structural models may benefit from incorporating more complex relations in their learning framework, such as the ones suggested here. Computing such relations could be expensive, and impractical for a large set of parts. A scheme based on interpretation of local units, which are then integrated and expanded as used in our model, will combine detailed interpretation with efficient computation.

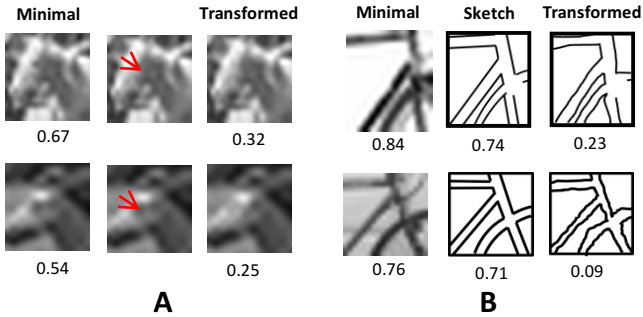

*Figure 3: 'Intervention': Testing informative relations via transformed minimal images. **(A).** Re-coloring a small set of pixels ( ≤ 4, pointed by the red arrow ) with the same color of their neighboring pixels. **(B). Rendering sketches from images**. In a transformed image, a relation is removed to test its predicted role in human perception. Relations tested include minimum intensity (in A), and high contour straightness (in B).*

|   | Relation Description |    | Relation Description |
|---|---|---|---|
| 1 | **Location and relative location:** for all primitives, and for all pairs of primitives in the structure. | 8 | **Length ratio** between two contours |
| 2 | **Strength of intensity maxima/minima**, center-surround filter responses at a point location. | 9 | **Parallelism** between two contours |
| 3 | **Deviation from line/circular arc:** in contours (particular for man-made objects). | 10 | **Coherent visual appearance** similar appearance/texture features in region i and in region j |
| 4 | **Visual appearance along contour** distribution of visual appearance/texture features (based on FCNN) along contour. | 11 | **Cover of a point by a contour:** if a contour i covers a point j. For 'cover' refer to [9]. |
| 5 | **Visual appearance inside a region** distribution of visual appearance/texture features (based on FCNN) in a region. | 12 | **Contour Bridging:** Testing whether two disconnected contour elements can be bridged (linked in the edge map). |
| 6 | **Relative location of contour endings:** between endings of two different contours | 13 | **Containment**: if point i is inside region j |
| 7 | **Continuity:** smooth continuation between two given contour endings. | 14 | **Contour ends in a region:** if a contour i ends in a region j. |

*Table 1: Relations that were found useful for modeling interpreation.*